\documentclass[11pt,a4paper]{volta_report}

\usepackage[authoryear,sort&compress,round]{natbib}

\usepackage{amsmath,amsfonts,bm}

\def\eqref#1{(\ref{#1})}
\def\1{\bm{1}}

\DeclareMathAlphabet{\mathsfit}{\encodingdefault}{\sfdefault}{m}{sl}
\SetMathAlphabet{\mathsfit}{bold}{\encodingdefault}{\sfdefault}{bx}{n}

\usepackage{graphicx}
\usepackage{float}
\usepackage{booktabs}
\usepackage{array}
\usepackage{amssymb}
\usepackage{flafter}
\usepackage{needspace}
\usepackage{enumitem}
\usepackage{placeins}
\usepackage{microtype}
\usepackage{hyperref}
\usepackage{xurl}
\hypersetup{
  hidelinks,
  pdftitle={ElectrolyteFM: Unifying Electrolyte Property Prediction through Cross-Property Knowledge Learning},
  pdfauthor={Jiaxin Yu, Shuo Wang, Peng Wang, Yongcai Wang, Deying Li}
}

\title{ElectrolyteFM: Unifying Electrolyte Property Prediction through Cross-Property Knowledge Learning}
\author{%
  \textbf{Jiaxin Yu}\textsuperscript{1,2},
  \textbf{Shuo Wang}\textsuperscript{1,2,\textdagger},
  \textbf{Peng Wang}\textsuperscript{1,2},
  \textbf{Yongcai Wang}\textsuperscript{2,*},
  \textbf{Deying Li}\textsuperscript{2}\\[0.45em]
  {\normalfont\fontsize{8}{10}\selectfont
  \textsuperscript{1}VoltaAI, Beijing, China\\
  \textsuperscript{2}Renmin University of China, Beijing, China\\
  \textsuperscript{*}Corresponding author:
  \href{mailto:ycw@ruc.edu.cn}{ycw@ruc.edu.cn}\\
  \textsuperscript{\textdagger}Project Leader, VoltaAI:
  \href{mailto:wangs@voltaai.cn}{wangs@voltaai.cn}}%
}
\runningtitle{ElectrolyteFM}
\reportlogo{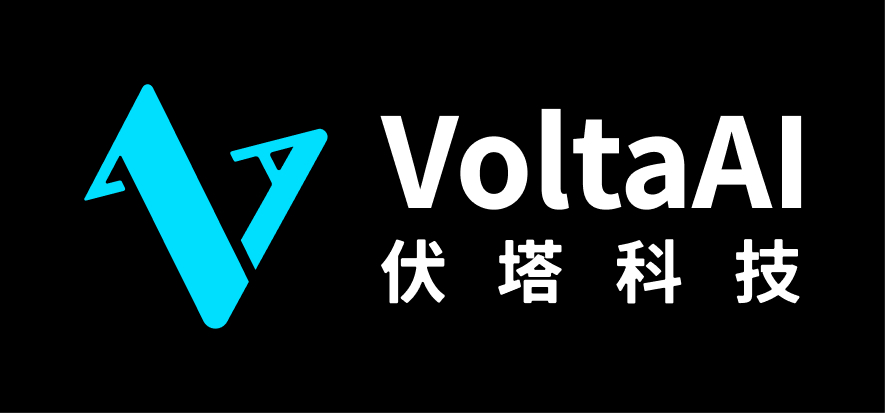}
\renewcommand{\today}{2026-09-30}

\setlist[itemize]{topsep=4pt,itemsep=3pt,parsep=0pt,partopsep=0pt}


\begin{document}
\maketitle

\begin{abstract}
Electrolyte formulation design requires balancing multiple physicochemical
properties, yet existing models often focus on a limited subset. Learning each
property in isolation can overlook transferable chemical information, whereas
indiscriminate sharing can introduce cross-property interference. Our directed
transfer analysis shows that jointly learning two property prediction tasks
can improve or degrade prediction relative to separate training, with
asymmetric transfer effects between the tasks. We propose ElectrolyteFM, a
unified multi-property prediction model which can more accurately predict
multiple properties of each electrolyte by effectively identifying and
utilizing property-specific features and knowledge shared across properties.
More specifically, ElectrolyteFM learns property-specific representations
independently and captures cross-property knowledge through a separately
trained expert pool. A router selects relevant shared information for each
formulation and target property, and property-specific residual adapters
convert this information into corrections to the corresponding representation
for prediction. Experiments on Electrolyte12 show that ElectrolyteFM reduces
normalized mean absolute error averaged across 12 electrolyte properties by
14.8\% relative to the strongest electrolyte-specific baseline. On an
independent sodium-electrolyte dataset unseen during training, it reduces
conductivity mean absolute error by 6.7\% relative to the best-performing
baseline.
\end{abstract}

\section{Introduction}
\label{sec:introduction}

Electrolytes govern ion transport, interfacial stability, and other coupled processes that determine the performance, lifetime, and safety of rechargeable batteries. Designing high-performing electrolyte formulations for practical battery applications is a multi-objective optimization problem over a combinatorial space of salts, solvents, compositions, and operating conditions, requiring a balance among multiple performance objectives. Deep-learning-based property prediction offers an efficient way to evaluate candidate formulations and guide the search toward those that meet application-specific performance requirements~\citep{zhang2024molsets,geomix2025,bamboomixer2026,wang2026scan}.

Existing electrolyte predictors often focus on a limited set of properties,
such as conductivity, providing incomplete information for formulation
selection~\citep{geomix2025,wang2026scan}. 
% Because different properties
% characterize the same formulation and arise from coupled chemical processes,
% predicting them jointly offers opportunities to share useful information.
% However, learning all properties through a common representation may introduce
% cross-property interference~\citep{liu2019negative_transfer}. To examine this
% issue, we compare single-property training with joint training of each
% property pair using a shared encoder. Across 12 properties,
% Figure~\ref{fig:directed-transfer-matrix}a reveals 61 positive and 71 negative
% directed transfer relations, with asymmetric effects between property pairs.
Electrolyte formulations are characterized not by a single property, but by a family of observables, including transport, solvation, thermodynamic, and electrochemical properties, which jointly determine their practical behavior. These properties arise from the same underlying formulation, molecular interactions, composition, and operating conditions, suggesting that they should not be modeled as isolated prediction problems. A unified predictive model that learns across heterogeneous electrolyte properties could reuse information across datasets, improve data efficiency, and provide a more comprehensive representation of electrolyte behavior.

\begin{figure}[!t]
    \centering
    \scalebox{0.80}{
        \begin{minipage}{\textwidth}
            \centering
            \includegraphics[width=0.70\textwidth]{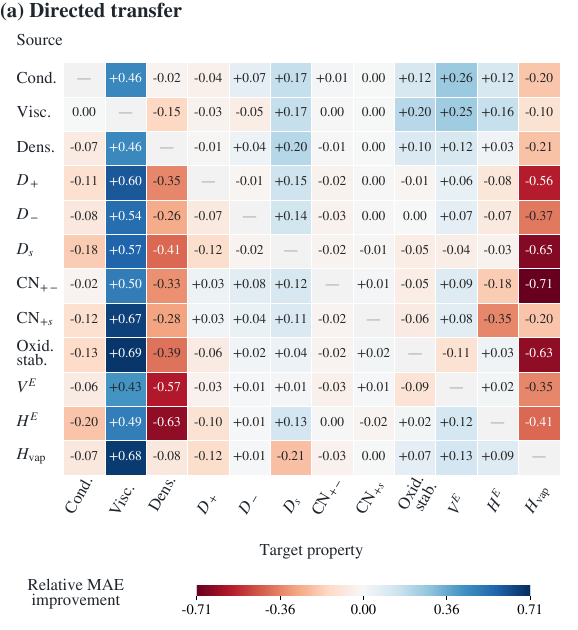}
            \hfill
            \includegraphics[width=0.28\textwidth]{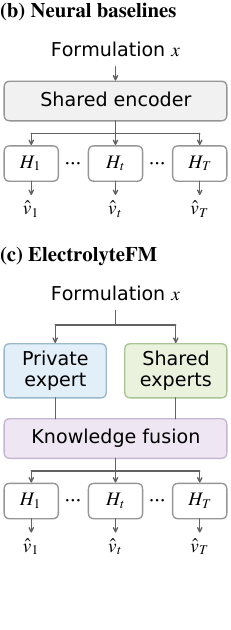}
        \end{minipage}
    }
    \caption{Cross-property transfer and sharing strategies. \textbf{(a)} Directed
    transfer scores on the test set used in Table~\ref{tab:main-mae}
    (rows: sources; columns: targets). Off-diagonal entries show relative target-MAE
    reductions; positive/negative values indicate beneficial
    transfer/interference. The score is defined in
    Section~\ref{sec:cross-property-sharing}.
    \textbf{(b)} Shared-representation prediction with property-specific heads.
    \textbf{(c)} ElectrolyteFM fuses private and shared representations
    before prediction through property-specific heads. Abbreviations: conductivity
    (Cond.), viscosity (Visc.), density (Dens.), oxidation stability
    (Oxid. stab.), and coordination number (CN).}
    \label{fig:directed-transfer-matrix}
\end{figure}

% These heterogeneous effects may reflect the different microscopic and
% macroscopic processes described by electrolyte properties, whose prediction
% can require different features and feature combinations. Consequently, even
% chemically related properties may not benefit equally from a common
% representation, which can hinder knowledge reuse or introduce interference.
However, a common physical origin does not imply uniform transferability across properties. Different observables reflect distinct physical processes and therefore may benefit differently from cross-property supervision.
To characterize this structure, we systematically measure directed transfer among 12 electrolyte properties by evaluating how supervision from each source property affects the prediction of every target property. As shown in Fig.~\ref{fig:directed-transfer-matrix}, 61 of the 132 directed transfer scores are positive and 71 are negative. Moreover, these effects are
strongly asymmetric: knowledge that benefits one property may provide little benefit, or even cause interference, in the reverse direction.

These observations motivate a model that preserves property-specific knowledge while selectively incorporating shared information. To this end, we propose \textbf{ElectrolyteFM}, which learns private and shared representations separately and then adapts shared knowledge to complement each property's private representation, enabling accurate prediction across a broad range of electrolyte properties.

Specifically, ElectrolyteFM adopts a \textbf{Shared--Specific Knowledge Learning} framework comprising three components. Specific Knowledge Learning (SpKL) captures property-specific predictive patterns with an independent private expert for each property. Shared Knowledge Learning (ShKL) learns reusable cross-property knowledge through a shared expert pool, with formulation--property conditioned routing selecting relevant shared information. Shared-to-Specific Knowledge Fusion (S2SKF) transforms the routed shared information into complementary corrections to the private representations through property-specific residual adapters.

\Needspace{4\baselineskip}
Our contributions are threefold:
\begin{itemize}
    \item \textbf{Directed Cross-Property Transfer Analysis.}
    We systematically characterize directed transfer among 12 electrolyte
    properties, revealing beneficial, harmful, and asymmetric interactions
    that provide an empirical basis for selective cross-property knowledge
    sharing.

    \item \textbf{Shared--Specific Knowledge Learning.}
    We propose ElectrolyteFM, which combines independently learned
    property-specific representations with shared representations from a
    separately trained expert pool through formulation--property conditioned
    routing and residual fusion.

    \item \textbf{Multi-property evaluation and external generalization.}
    We curate Electrolyte12 and evaluate multi-property prediction, learning strategies, 
    and knowledge-branch contributions, together with zero-shot generalization 
    and few-shot adaptation on an independent sodium-electrolyte dataset.
\end{itemize}

Across 12 electrolyte properties, ElectrolyteFM reduces macro normalized
mean absolute error (NMAE) by 14.8\% relative to the strongest
electrolyte-specific baseline. On an independent sodium-electrolyte dataset,
it reduces conductivity MAE by 6.7\% relative to the best-performing baseline
evaluated on that dataset, with further gains from few-shot adaptation.

% We propose ElectrolyteFM, which decouples Private Learning and Shared
% Learning to preserve an independent predictive representation for each
% property. After learning both branches separately, we freeze them and
% train property-specific residual adapters during Adaptive Fusion to
% selectively incorporate shared information into each private predictor.
\section{Related Work}
\label{sec:related_work}

\subsection{Learning for Electrolyte Formulations}

Learning-based electrolyte modeling builds on molecular pretraining and
formulation-level representations. Molecular pretraining methods, including
MolT5 and Uni-Mol, learn transferable representations from molecular strings,
graphs, language, and three-dimensional
structures~\citep{edwards2022molt5,xia2023molebert,zhou2023unimol,ji2024unimol2}.
At the formulation level, MolSets, GeoMix, and SCAN explicitly model mixture
composition and component interactions for electrolyte property
prediction~\citep{zhang2024molsets,geomix2025,wang2026scan}. DiffMix and related
mixture models incorporate thermodynamic structure into property
prediction~\citep{diffmix2024,specht2024hanna}, while BAMBOO connects microscopic
interactions with macroscopic electrolyte properties through learned interatomic
potentials and molecular dynamics~\citep{bamboo2025}. Uni-ELF and Bamboo-Mixer
further explore formulation-level pretraining and unified predictive--generative
modeling, respectively, to support property prediction and formulation
design~\citep{unielf2024,bamboomixer2026}. MixUni combines multi-property parameter sharing with a physics-informed
conductivity head for liquid mixtures~\citep{huang2026mixuni}.

Joint learning can benefit some targets while interfering with others.
ElectrolyteFM learns property-specific and shared representations separately,
then uses property-specific residual adapters to incorporate shared information
into the private representations.

\subsection{Cross-Task Transfer Learning}

Cross-task transfer learning seeks to reuse knowledge across prediction tasks,
but sharing can also introduce negative transfer~\citep{liu2019negative_transfer}.
Taskonomy characterizes directed transfer between visual tasks, while subsequent
multi-task studies examine cooperation and competition under joint
learning~\citep{zamir2018taskonomy,standley2020tasks}. Task affinities and predicted
task-combination gains further guide task grouping and selective group
updates~\citep{fifty2021taskgroupings,song2022multitaskgrouping,selectivetaskgrouping2025}.
At the optimization level, PCGrad and CAGrad address conflicting task gradients,
while Nash-MTL and FAMO balance task contributions through bargaining-based
aggregation and adaptive loss weighting,
respectively~\citep{pcgrad2020,cagrad2021,navon2022nashmtl,liu2023famo}.
In scientific prediction, cross-property transfer and source-selection strategies
exploit data-rich properties and estimated task similarities to support
data-scarce targets~\citep{crossproperty2021,motse2022,transferabilitymap2024}.
Adaptive checkpointing and specialization further address negative transfer in
molecular property learning~\citep{molecularmtl2025}.

Task selection and grouping determine which tasks share supervision, while
gradient coordination regulates joint optimization. These mechanisms do not
directly specify how shared knowledge contributes to an individual
formulation--property prediction. Motivated by directed transfer analysis,
ElectrolyteFM conditions expert allocation on both the formulation and target
property to select relevant shared information.

\subsection{Mixture-of-Experts for Knowledge Sharing}

Mixture-of-experts architectures support selective knowledge sharing through
learned combinations of expert representations. MMoE uses task-specific gates
over shared experts, PLE explicitly separates shared and task-specific experts,
and TaskExpert dynamically assembles task-specific features from multiple expert
representations~\citep{mmoe2018,ple2020,ye2023taskexpert}. Related architectures
regulate sharing through selective layer specialization, orthogonal expert
representations, or modality-specific model
spaces~\citep{shi2023recon,hendawy2024moore,peng2025sm4}. Another line of work reuses
pretrained task modules: PEMT and MeteoRA selectively combine task-specific
adapters in language models~\citep{lin2024pemt,xu2025meteora}, while materials
expert-composition methods and MoMa combine source-property experts or
specialized modules for downstream prediction~\citep{chang2022materialsexperts,moma2026}.

MMoE and PLE control expert allocation through routing, but task-specific
representations remain coupled to joint optimization. Source-expert
composition instead adapts separately pretrained modules to downstream
targets. ElectrolyteFM learns property-specific private representations and
a shared expert pool separately, then freezes both branches 
for fusion.
Property-specific residual adapters convert routed shared information into
corrections to the private representations.

\begin{figure}[ht]
    \centering
    \includegraphics[width=0.9\textwidth]{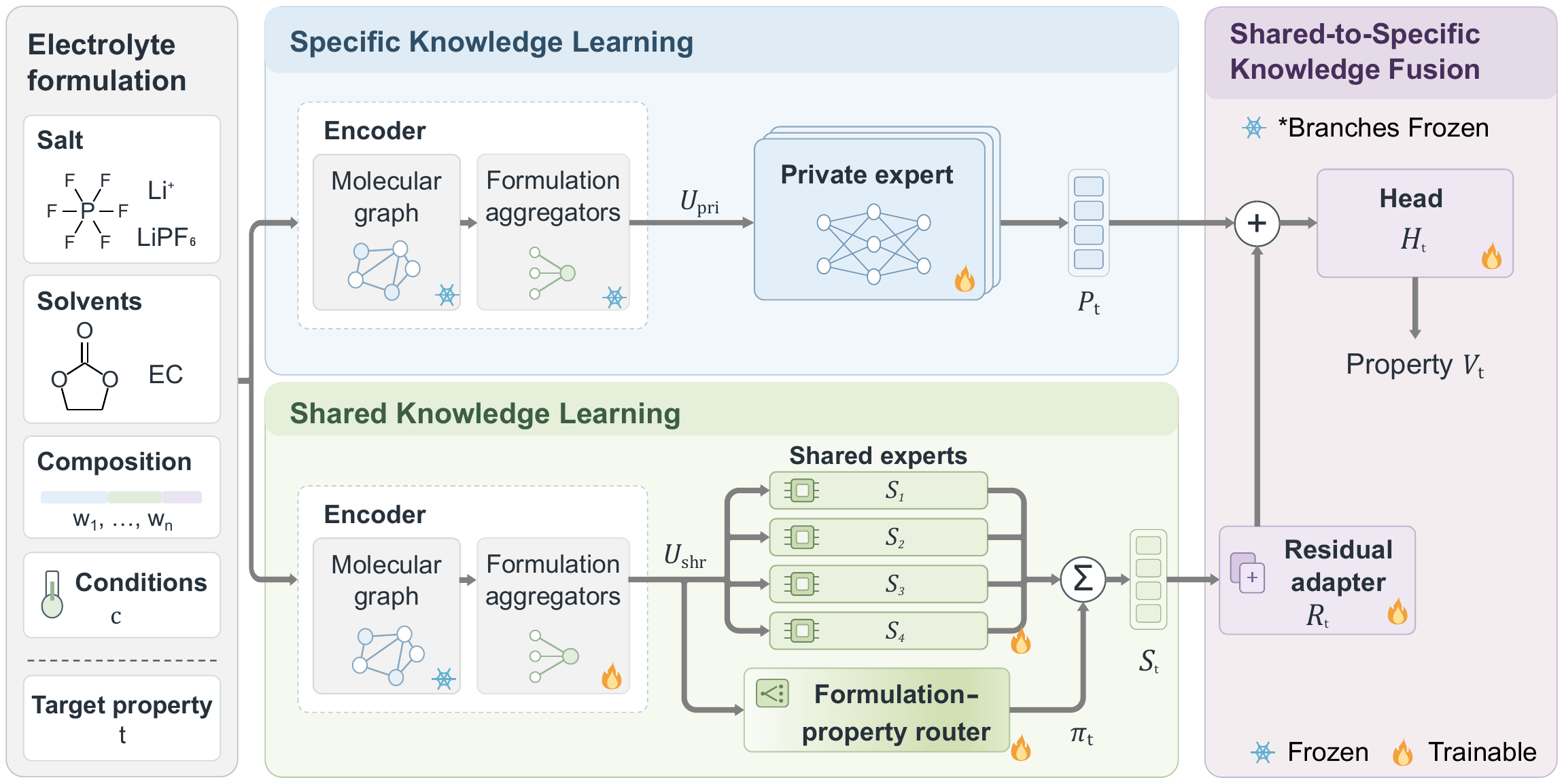}
    \caption{Overview of ElectrolyteFM. In Specific Knowledge Learning (SpKL),
    private experts learn property-specific knowledge. In Shared Knowledge
    Learning (ShKL), a shared expert pool learns reusable cross-property knowledge,
    and formulation--property conditioned routing combines shared expert
    outputs according to the input formulation and target property.
    Shared-to-Specific Knowledge Fusion uses property-specific residual
    adapters to incorporate the routed shared information into private
    representations, with both learned branches frozen. Snowflakes and
    flames indicate frozen and trainable modules, respectively.}
    \label{fig:method-overview}
\end{figure}
% Place the overview figure before Method without forcing a new page.

\section{Method}
\label{sec:method}

\subsection{Directed Cross-Property Transfer Analysis}
\label{sec:cross-property-sharing}

We first characterize how supervision from one electrolyte property affects the prediction of another. For each ordered source--target pair $(a,t)$, we compare a target-only model trained on property $t$ with a pairwise model jointly trained on properties $a$ and $t$. Both models use the same architecture and initialization, consisting of a shared backbone and property-specific prediction heads, as illustrated in Figure~\ref{fig:directed-transfer-matrix}(b).

To quantify the directional effect of source property $a$ on target property $t$, we define the directed transfer score as the relative reduction in target mean absolute error (MAE): 
\begin{equation} 
\Delta_{a\rightarrow t} = \frac{ \mathrm{MAE}^{\mathrm{single}}_t - \mathrm{MAE}^{\mathrm{pair}(a,t)}_t }{ \mathrm{MAE}^{\mathrm{single}}_t }. 
\label{eq:directed-transfer} 
\end{equation} 
Both models are evaluated on the same test samples of the target property, with MAE computed in physical units. A positive $\Delta_{a\rightarrow t}$ indicates that supervision from property $a$ improves prediction of property $t$, whereas a negative value indicates cross-property interference.

As shown in Figure~\ref{fig:directed-transfer-matrix}(a), cross-property transfer is highly heterogeneous and directional: the same source property can benefit some targets while interfering with others, and the transfer effect between two properties is generally asymmetric. For example, adding conductivity supervision reduces solvent diffusivity MAE by 17.2\%, whereas adding solvent diffusivity supervision increases conductivity MAE by 17.9\%. These observations motivate ElectrolyteFM to preserve independent property-specific representations while selectively incorporating shared knowledge according to the target property and input formulation.

% \subsection{Problem Formulation and Overview}
% \label{sec:problem-formulation}
% \label{sec:framework-overview}

% Let $\mathcal{T}=\{1,\ldots,T\}$ denote the electrolyte properties, with
% $T=12$ in our study. Each formulation $x_i$ specifies its molecular
% components, composition, and experimental conditions. Since labels are
% partially observed, each property has a labeled set
% \[
%     \mathcal{D}_t=\{(x_i,y_{it})\mid m_{it}=1\},
% \]
% where $m_{it}$ indicates label availability. Our goal is to learn a
% predictor $\widehat y_{it}=f(x_i,t)$ for a given formulation and target
% property.

% Figure~\ref{fig:method-overview} summarizes ElectrolyteFM. We adopt the
% molecular and formulation encoders of
% Bamboo-Mixer~\citep{bamboomixer2026}, with separate copies producing
% $\mathbf{u}_i^{\mathrm{pri}},\mathbf{u}_i^{\mathrm{shr}}\in\mathbb{R}^{d_u}$
% for the two branches. The private expert $P_t$ produces a property-specific
% representation $\mathbf{p}_{it}$, while the shared experts
% $\{S_k\}_{k=1}^{K}$ are combined into $\mathbf{s}_{it}$ using routing
% weights $\boldsymbol{\pi}_{it}$ conditioned on the formulation and target
% property. A residual adapter $R_t$ maps $\mathbf{s}_{it}$ into a correction
% to $\mathbf{p}_{it}$, and the final head $H_t$ outputs the standardized
% prediction $\widehat v_{it}$.

% We use mean squared error on transformed, standardized targets $v_{it}$
% (z-MSE) for the learning objectives.

\subsection{Problem Formulation and Overview}
\label{sec:problem-formulation}
\label{sec:framework-overview}

Let $\mathcal{T}=\{1,\ldots,T\}$ denote the set of electrolyte properties,
with $T=12$ in this work. Each formulation $x_i$ is defined by its molecular
components, composition, and experimental conditions. Since property labels
are partially observed, the supervision for property $t$ is
\begin{equation}
    \mathcal{D}_t=\{(x_i,y_{it})\mid m_{it}=1\},
\end{equation}
where $m_{it}$ indicates label availability. Our goal is to learn a unified
predictor
\begin{equation}
    \widehat{y}_{it}=f(x_i,t)
\end{equation}
that captures both property-specific and cross-property knowledge.

Figure~\ref{fig:method-overview} illustrates ElectrolyteFM. Separate copies
of the molecular and formulation encoders produce private and shared
representations $\mathbf{u}_i^{\mathrm{pri}}$ and
$\mathbf{u}_i^{\mathrm{shr}}$. Specific Knowledge Learning (SpKL) uses a
property-specific expert $P_t$ to obtain $\mathbf{p}_{it}$, while Shared
Knowledge Learning (ShKL) combines shared expert outputs using
formulation--property conditioned routing weights $\boldsymbol{\pi}_{it}$
to obtain $\mathbf{s}_{it}$. Shared-to-Specific Knowledge Fusion (S2SKF)
then uses a property-specific residual adapter $R_t$ to convert
$\mathbf{s}_{it}$ into a correction to $\mathbf{p}_{it}$, followed by the
prediction head $H_t$.

All objectives are defined on transformed and standardized targets $v_{it}$,
using mean squared error in the standardized space (\emph{z-MSE}).

\subsection{Specific Knowledge Learning}
\label{sec:private-learning}

Specific Knowledge Learning captures property-specific predictive patterns through independent experts. For a query $(x_i,t)$, the corresponding
private expert $P_t$ transforms $\mathbf{u}^{\mathrm{pri}}_i$ into a private
state $\mathbf{p}_{it}$, and its head $H^{\mathrm{pri}}_t$ produces the
standardized prediction:
\begin{equation}
    \mathbf{p}_{it}=P_t(\mathbf{u}^{\mathrm{pri}}_i),
    \qquad
    \widehat v^{\mathrm{pri}}_{it}
        =H^{\mathrm{pri}}_t(\mathbf{p}_{it}).
\label{eq:private-branch}
\end{equation}
Each $P_t$ is a multilayer perceptron (MLP) producing a private state
$\mathbf{p}_{it}\in\mathbb{R}^{d_h}$, where $d_h$ is the hidden dimension.
Each property has its own expert and scalar prediction head.

We train each expert and its head only on $\mathcal{D}_t$, minimizing
$\mathcal{L}^{\mathrm{pri}}_t
=\mathbb{E}_{\mathcal{D}_t}[\ell(\widehat v^{\mathrm{pri}}_{it},v_{it})]$,
with the complete encoder frozen.

\subsection{Shared Knowledge Learning}
\label{sec:shared-learning}

Shared Knowledge Learning captures transferable patterns across properties using a pool of $K$ shared experts $\{S_k\}_{k=1}^{K}$. To account for the heterogeneous relevance of shared knowledge, we introduce a formulation--property conditioned router:
\begin{equation}
    \boldsymbol{\pi}_{it}
       = \operatorname{softmax}\!\left(
           W_r\mathbf{u}^{\mathrm{shr}}_i+\mathbf{b}_t
         \right),
\label{eq:mixed-router}
\end{equation}
where $W_r\in\mathbb{R}^{K\times d_u}$ is a bias-free linear projection.
The learned vector $\mathbf{b}_t\in\mathbb{R}^{K}$ expresses the property's
preference over experts, while $W_r\mathbf{u}^{\mathrm{shr}}_i$ adjusts that
preference for the input formulation. Each expert receives the same shared
input, and its contribution is weighted by the corresponding router
output:
\begin{equation}
    \mathbf{s}_{it}
       = \sum_{k=1}^{K}\pi_{itk}S_k(\mathbf{u}^{\mathrm{shr}}_i),
    \qquad
    \widehat v^{\mathrm{shr}}_{it}
       = H^{\mathrm{shr}}_t(\mathbf{s}_{it}).
\label{eq:shared-branch}
\end{equation}
A property-specific ShKL prediction head $H^{\mathrm{shr}}_t$ maps the
routed shared representation $\mathbf{s}_{it}\in\mathbb{R}^{d_h}$ to the
standardized prediction and provides supervision to the shared experts
and router.

ShKL draws a property with probability
$q(t)\propto|\mathcal{D}_t|^{\alpha}$, with $0<\alpha<1$ to moderate
the influence of data-rich properties, and then samples a batch from its labeled
set, optimizing
\begin{equation}
    \mathcal{L}^{\mathrm{shr}}
      = \mathbb{E}_{t\sim q}
         \mathbb{E}_{(x_i,y_{it})\sim\mathcal{D}_t}
         \left[\ell(\widehat v^{\mathrm{shr}}_{it},v_{it})\right].
\label{eq:shared-loss}
\end{equation}
The neutral and ionic formulation aggregators, shared experts, router, and
ShKL prediction heads are trained jointly, with the molecular graph module
fixed. We retain the learned shared encoder, experts, and router to
produce $\mathbf{s}_{it}$ for S2SKF. These prediction heads are used only
during ShKL.

\subsection{Shared-to-Specific Knowledge Fusion}
\label{sec:fusion}

Shared-to-Specific Knowledge Fusion (S2SKF) integrates transferable shared
knowledge into the independently learned specific representation. For each
property $t$, a property-specific residual adapter $R_t$ transforms the routed
shared representation $\mathbf{s}_{it}$ into a complementary residual:
\begin{equation}
    \mathbf{z}_{it}
        =
        \mathbf{p}_{it}
        +
        R_t(\mathbf{s}_{it}),
    \qquad
    \widehat v_{it}
        =
        H_t(\mathbf{z}_{it}),
\label{eq:fusion}
\end{equation}
where $\mathbf{p}_{it}$ denotes the specific representation learned for
property $t$, and $\mathbf{z}_{it}$ is the fused representation.

The adapter adopts a bottleneck architecture:
\begin{equation}
    R_t(\mathbf{s}_{it})
      =
      W_t^{\mathrm{up}}
      \operatorname{SiLU}\!\left(
          W_t^{\mathrm{down}}\operatorname{LN}(\mathbf{s}_{it})
          +\mathbf{a}_t^{\mathrm{down}}
      \right)
      +\mathbf{a}_t^{\mathrm{up}},
\label{eq:adapter}
\end{equation}
where $d_h$ is the representation dimension and $d_b<d_h$ is the bottleneck dimension.

We zero-initialize the up-projection $W_t^{\mathrm{up}}$ and
$\mathbf{a}_t^{\mathrm{up}}$, and initialize $H_t$ from the corresponding
specific prediction head. The fused model therefore starts from the learned
specific predictor and gradually incorporates useful shared information.
During fusion, the specific and shared branches are frozen, and only $R_t$
and $H_t$ are optimized:
\begin{equation}
    \mathcal{L}^{\mathrm{fus}}_t
    =
    \mathbb{E}_{(x_i,y_{it})\sim\mathcal{D}_t}
    \left[
        \ell(\widehat v_{it},v_{it})
    \right].
\label{eq:fusion-loss}
\end{equation}

This formulation preserves property-specific predictive knowledge while
allowing transferable cross-property information to selectively complement
it. 

\section{Experiments}
\label{sec:experiments}

\subsection{Experimental Setup}
\label{sec:experimental-settings}

\paragraph{Datasets and settings.}
We evaluate all methods on \textbf{Electrolyte12}, a large-scale electrolyte dataset curated in this work, covering 12 electrolyte properties.
The training and validation sets contain both experimental and computed labels, comprising 366{,}825 and 44{,}811 records, respectively.
The test set contains 43{,}709 records, including experimental measurements for conductivity, viscosity, excess molar enthalpy, and excess molar volume, and computed labels for the remaining eight properties.
All methods use the same training, validation, and test splits.

\paragraph{Baselines.}
We compare ElectrolyteFM with RDKit-XGB~\citep{landrum2024rdkit,chen2016xgboost}, MolT5-XGB~\citep{edwards2022molt5,chen2016xgboost},
Bamboo-Mixer~\citep{bamboomixer2026}, GeoMix~\citep{geomix2025},
MolSets~\citep{zhang2024molsets}, and SCAN~\citep{wang2026scan}.
All methods follow the same data and target preprocessing, including target transformations and missing-label handling.

\paragraph{Evaluation metrics.}
We use mean absolute error (MAE) as the primary metric for evaluating individual electrolyte properties, with errors reported in their original physical units.
To summarize performance across properties with different numerical scales, we additionally report macro normalized MAE (NMAE), defined as
$\frac{1}{|G|}\sum_{t\in G}\mathrm{MAE}_t/\sigma^{\mathrm{train}}_t$,
where $G$ denotes the set of evaluated properties and $\sigma^{\mathrm{train}}_t$ is the standard deviation of property $t$ in the training set.
We further report Pearson correlation coefficient (PCC) and coefficient of determination ($R^2$). 

\subsection{Multi-Property Prediction}
\label{sec:overall-results}

Table~\ref{tab:main-mae} compares the methods on Electrolyte12 in terms of per-property MAE, macro NMAE across all twelve properties, and the number of properties on which each method achieves the best MAE. Results are grouped according to whether the test labels are obtained experimentally or computationally.

\begin{table}[!htbp]
    \centering
    \fontsize{8}{9.4}\selectfont
    \setlength{\tabcolsep}{1.6pt}
    \setlength{\arrayrulewidth}{0.3pt}
    \renewcommand{\arraystretch}{1.08}
    \newcommand{\maehead}[3]{%
        \begingroup\fontsize{8}{9}\selectfont
        \begin{tabular}[c]{@{}c@{}}#1\\[-1pt]
        {\fontsize{7}{8}\selectfont #2}\\[-1pt]
        {\fontsize{7}{8}\selectfont #3}\end{tabular}\endgroup}
    \caption{Per-property MAE on Electrolyte12 (lower is better).
    Blocks denote experimental and computed test labels. Bold and underline
    indicate the best and second-best unrounded values. The macro and
    best-count rows summarize all twelve properties.}
    \label{tab:main-mae}
    \scalebox{0.9}{
    \begin{tabular*}{\textwidth}{@{}l|@{\extracolsep{\fill}}*{7}{c}@{}}
        \toprule
        {\fontsize{8}{9}\selectfont Property} &
        \maehead{RDKit-XGB}{(Landrum et al.}{\citeyear{landrum2024rdkit})} &
        \maehead{MolT5-XGB}{(Edwards et al.}{\citeyear{edwards2022molt5})} &
        \maehead{Bamboo-Mixer}{(Yang et al.}{\citeyear{bamboomixer2026})} &
        \maehead{GeoMix}{(Li et al.}{\citeyear{geomix2025})} &
        \maehead{MolSets}{(Zhang et al.}{\citeyear{zhang2024molsets})} &
        \maehead{SCAN}{(Wang and}{You \citeyear{wang2026scan})} &
        {\fontsize{8}{9}\selectfont\begin{tabular}[c]{@{}c@{}}\textbf{ElectrolyteFM}\\[-1pt]\textbf{(Ours)}\end{tabular}} \\
        \midrule
        \multicolumn{8}{l}{\textbf{Experimental labels}} \\
        Cond. (mS/cm) & 4.444 & 4.211 & \underline{1.766} & 3.211 & 1.980 & 2.335 & \textbf{0.881} \\
        Visc. (mPa$\cdot$s) & 156.457 & 156.654 & \underline{125.313} & 147.831 & 139.246 & 148.301 & \textbf{91.187} \\
        $H^{E}$ (J/mol) & 311.071 & 299.599 & \underline{167.964} & 374.825 & 191.577 & 248.297 & \textbf{54.092} \\
        $V^{E}$ (cm$^3$/mol) & 0.140 & 0.136 & \underline{0.068} & 0.181 & 0.111 & 0.142 & \textbf{0.050} \\
        \midrule
        \multicolumn{8}{l}{\textbf{Computed labels}} \\
        Dens. (g/cm$^3$) & 0.0566 & 0.0506 & \underline{0.0079} & 0.0296 & 0.0175 & 0.0321 & \textbf{0.0054} \\
        $D_{+}$ ($10^{-11}$ m$^2$/s) & 8.911 & 8.204 & 4.295 & 4.659 & \textbf{4.102} & \underline{4.114} & 4.144 \\
        $D_{-}$ ($10^{-11}$ m$^2$/s) & 8.785 & 7.806 & 3.897 & 4.095 & 3.846 & \underline{3.769} & \textbf{3.769} \\
        $D_{s}$ ($10^{-11}$ m$^2$/s) & 22.227 & 19.611 & 10.745 & 9.968 & \underline{9.130} & 9.949 & \textbf{8.843} \\
        $\mathrm{CN}_{+-}$ & 1.127 & 1.079 & \underline{0.948} & 0.949 & 0.962 & 0.966 & \textbf{0.946} \\
        $\mathrm{CN}_{+s}$ & 2.084 & 2.066 & 1.851 & \underline{1.799} & 1.885 & 1.912 & \textbf{1.788} \\
        Oxid. stab. (eV) & 0.517 & 0.555 & \textbf{0.210} & 0.638 & 0.389 & 0.518 & \underline{0.217} \\
        $H_{\mathrm{vap}}$ (J/mol) & 4242.9 & 4790.6 & \underline{684.9} & 2735.9 & 2569.0 & 2409.8 & \textbf{477.2} \\
        \midrule
        Macro NMAE & 0.3643 & 0.3536 & \underline{0.1845} & 0.2936 & 0.2279 & 0.2592 & \textbf{0.1571} \\
        \textbf{Best count} & 0/12 & 0/12 & \underline{1/12} & 0/12 & \underline{1/12} & 0/12 & \textbf{10/12} \\
        \bottomrule         
    \end{tabular*}}
\end{table}

ElectrolyteFM achieves the best MAE on 10 of the 12 properties, including all four properties with experimental labels and six of the eight properties with computed labels. It also obtains the lowest macro NMAE of 0.1571, compared with 0.1845 for the strongest baseline. The improvements are particularly pronounced on the experimentally measured properties: relative to the strongest baseline for each property, ElectrolyteFM reduces MAE by 50.1\% for conductivity, 27.2\% for viscosity, 67.8\% for excess molar enthalpy, and 26.7\% for excess molar volume. Among the computed properties, it further reduces MAE by 31.3\% for density and 30.3\% for heat of vaporization. 

\subsection{Generalization to Unseen Formulations}
\label{sec:ood-generalization}

We further evaluate generalization on an independent experimental dataset of 106 multicomponent sodium-electrolyte formulations, including mixtures with up to 14 components~\citep{munozperales2026sodium}. 

We evaluate conductivity prediction under two settings: zero-shot generalization, where the trained model is directly applied to the unseen sodium formulations, and few-shot adaptation, where a small number of target-domain labels are provided for adaptation.

\subsubsection{Zero-shot Generalization}
\label{sec:zero-shot}

Table~\ref{tab:conductivity-ood} compares conductivity prediction on the
experimental in-distribution (ID) subset and the independent
out-of-distribution (OOD) dataset.

\begin{table}[!htbp]
    \centering\small
    \setlength{\tabcolsep}{4pt}
    \caption{Conductivity prediction on experimental ID and independent OOD data. MAE is in mS/cm.}
    \label{tab:conductivity-ood}
    \scalebox{0.9}{
    \begin{tabular}{lrrrrrr}
        \toprule
        & \multicolumn{3}{c}{\shortstack{Experimental ID\\$N=7{,}153$}} & \multicolumn{3}{c}{\shortstack{OOD\\$N=106$}} \\
        \cmidrule(lr){2-4}\cmidrule(lr){5-7}
        Method & MAE $\downarrow$ & PCC $\uparrow$ & $R^2$ $\uparrow$ & MAE $\downarrow$ & PCC $\uparrow$ & $R^2$ $\uparrow$ \\
        \midrule
        SCAN & 2.335 & 0.856 & 0.723 & \underline{2.778} & \underline{0.775} & \underline{0.599} \\
        MolSets & 1.980 & 0.899 & 0.795 & 7.369 & 0.201 & -1.839 \\
        GeoMix & 3.211 & 0.771 & 0.574 & 3.321 & 0.685 & 0.208 \\
        Bamboo-Mixer & \underline{1.766} & \underline{0.909} & \underline{0.822} & 3.131 & 0.582 & 0.295 \\
        RDKit-XGB & 4.444 & 0.729 & 0.274 & 3.624 & 0.349 & -0.228 \\
        MolT5-XGB & 4.211 & 0.748 & 0.303 & 3.317 & 0.733 & 0.041 \\
        \textbf{ElectrolyteFM (Ours)} & \textbf{0.881} & \textbf{0.942} & \textbf{0.887} & \textbf{2.592} & \textbf{0.791} & \textbf{0.605} \\
        \bottomrule
    \end{tabular}}
\end{table}

The baseline ranking changes under distribution shift: Bamboo-Mixer
outperforms SCAN on the ID test set but falls behind it on the OOD test
set. Higher ID accuracy therefore does not necessarily translate into
better prediction on unseen formulations. ElectrolyteFM ranks first
across all three metrics in both settings, reducing OOD MAE by 6.7\%
relative to SCAN, the strongest OOD baseline. Its predictive advantage
thus extends to the independent sodium-electrolyte dataset without
target-domain adaptation.

\subsubsection{Few-shot Adaptation}
\label{sec:few-shot}

\begin{figure}[!htbp]
    \centering
    \includegraphics[width=\textwidth]{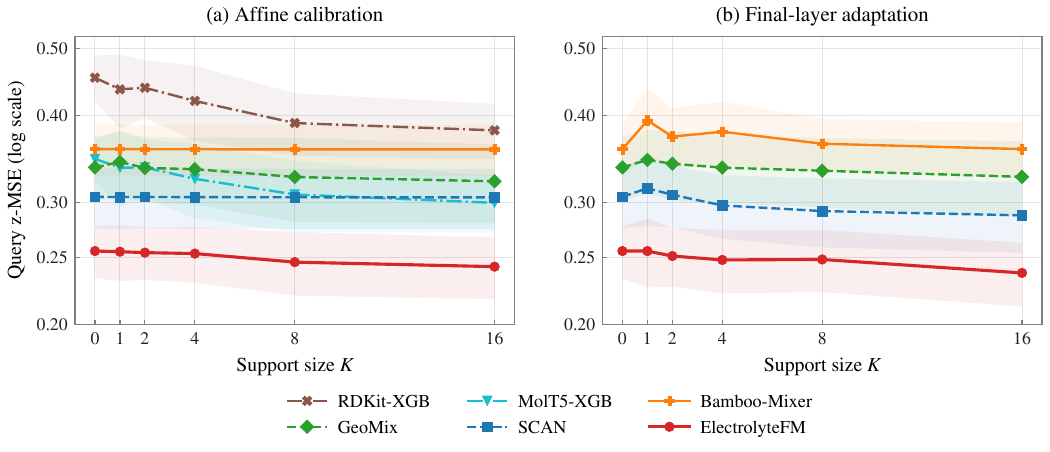}
    \caption{Conductivity OOD few-shot adaptation for six methods from
    Table~\ref{tab:main-mae}. Curves show mean query z-MSE on logarithmic y-axes; shaded
    regions show the sample standard deviation over 20 fixed episodes.
    XGBoost models are evaluated under affine calibration only.
    MolSets is omitted because some values exceed the plotted y-axis range.}
    \label{fig:ood-few-shot}
\end{figure}

We evaluate affine calibration and final-layer adaptation over 20 fixed
support/query episodes with nested support sizes $K\in\{0,1,2,4,8,16\}$.
Figure~\ref{fig:ood-few-shot} compares six methods from
Table~\ref{tab:main-mae} under affine calibration and four neural methods
under final-layer adaptation using query z-MSE. Each episode uses 74 query
formulations, held fixed across support sizes.

ElectrolyteFM achieves the lowest mean query z-MSE at every evaluated
support size under both protocols. With 16 labeled formulations and the
learned representations fixed, affine calibration and final-layer
adaptation reduce its mean query z-MSE from 0.2552 to 0.2423 and 0.2373,
respectively.

\subsection{Ablation Studies}
\label{sec:ablations}

\subsubsection{Comparison of Learning Strategies}
\label{sec:learning-strategies}

To assess how different learning strategies affect multi-property prediction,
we compare ElectrolyteFM with single-task learning (STL), multi-gate
mixture-of-experts (MMoE)~\citep{mmoe2018}, and progressive layered extraction
(PLE)~\citep{ple2020}. Table~\ref{tab:learning-strategies} reports the macro
metrics across all twelve properties, with equal weight assigned to each
property. All methods start
from the same pretrained molecular encoders. STL learns each property
independently, whereas MMoE and PLE learn jointly across properties;
ElectrolyteFM separates representation learning from subsequent
target-specific fusion.

Both multi-task baselines outperform STL on all three macro metrics,
demonstrating the benefit of cross-property learning in this comparison.
However, the label-source breakdown in Table~\ref{tab:learning-strategies-full}
shows that MMoE and PLE improve all three macro metrics over STL for the
experimental-label group, but worsen them for the computed-label group.
ElectrolyteFM improves all three metrics over STL in both groups.
ElectrolyteFM further reduces macro NMAE by 2.5\% relative to PLE while
achieving the highest macro PCC and $R^2$. ElectrolyteFM combines decoupled
knowledge learning with shared-to-specific fusion, using shared information
to complement independently learned property-specific representations.
This combination achieves better aggregate predictive performance than
either independent learning or the two joint-learning alternatives.

\begin{table}[!htbp]
    \centering
    \begin{minipage}[t]{0.49\textwidth}
        \centering
        \caption{Learning strategies: macro metrics over all 12 properties.}
        \label{tab:learning-strategies}
        \small
        \setlength{\tabcolsep}{2pt}
        \begin{tabular*}{\linewidth}{@{}l@{\extracolsep{\fill}}rrr@{}}
            \toprule
            Method & NMAE $\downarrow$ & PCC $\uparrow$ & $R^2$ $\uparrow$ \\
            \midrule
            STL & 0.1653 & 0.8931 & 0.8071 \\
            MMoE & 0.1641 & 0.9058 & 0.8220 \\
            PLE & \underline{0.1611} & \underline{0.9228} & \underline{0.8228} \\
            \textbf{ElectrolyteFM} & \textbf{0.1571} & \textbf{0.9268} & \textbf{0.8357} \\
            \bottomrule
        \end{tabular*}
    \end{minipage}\hfill
    \begin{minipage}[t]{0.47\textwidth}
        \centering
        \caption{Knowledge learning and fusion ablation: macro NMAE over all 12 properties.}
        \label{tab:stage-ablation}
        \small
        \setlength{\tabcolsep}{2pt}
        \begin{tabular*}{\linewidth}{@{}c@{\extracolsep{\fill}}ccr@{}}
            \toprule
            SpKL & ShKL & S2SKF & NMAE $\downarrow$ \\
            \midrule
            $\checkmark$ & -- & -- & 0.1841 \\
            -- & $\checkmark$ & -- & \underline{0.1674} \\
            $\checkmark$ & $\checkmark$ & $\checkmark$ & \textbf{0.1571} \\
            \bottomrule
        \end{tabular*}
    \end{minipage}
\end{table}

\subsubsection{Contribution of Knowledge Learning and Fusion}
\label{sec:stage-ablation}

We compare the private branch learned through SpKL, the shared branch
learned through ShKL, and their combination through S2SKF.
Table~\ref{tab:stage-ablation} reports macro NMAE over all twelve properties.
Unlike STL,
SpKL keeps the formulation aggregators fixed.
The shared branch alone outperforms the private branch, while the complete
model reduces macro NMAE by 14.7\% and 6.1\% relative to the private and
shared branches, respectively.

The improvement over the shared branch alone indicates that the
independently learned private representation provides complementary
predictive information. S2SKF uses this private representation as a
foundation and learns a property-specific residual correction from the
routed shared representation. The lower macro NMAE relative to either branch alone supports
combining property-specific and shared knowledge through shared-to-specific
fusion.

\subsubsection{Expert Allocation and Predictive Contributions}
\label{sec:expert-analysis}

\begin{figure}[!htbp]
    \centering
    \begin{minipage}[b]{0.57\textwidth}
        \centering
        \includegraphics[height=0.27\textheight]{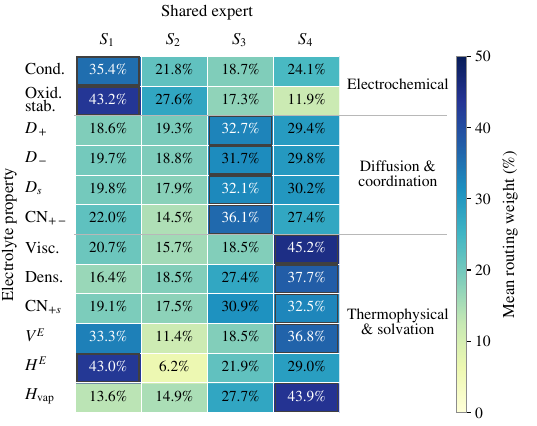}
        \par\smallskip
        (a)
    \end{minipage}\hspace{0.035\textwidth}%
    \begin{minipage}[b]{0.29\textwidth}
        \centering
        \includegraphics[height=0.285\textheight]{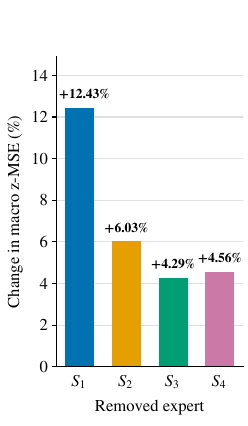}
        \par\smallskip
        (b)
    \end{minipage}
    \caption{Expert allocation and contribution in ElectrolyteFM.
    (a) Mean routing weights by property; boxes mark each row maximum.
    (b) Relative increase in macro z-MSE after removing each expert at
    inference and renormalizing the remaining weights. All twelve
    properties receive equal weight.}
    \label{fig:routing-mechanism}
\end{figure}

Figure~\ref{fig:routing-mechanism}(a) shows mean routing weights by
property. $S_1$ receives the largest weight for conductivity, $H^{E}$, and
oxidation stability; $S_3$ for $D_{+}$, $D_{-}$, $D_s$, and
$\mathrm{CN}_{+-}$; and $S_4$ for viscosity, $V^{E}$, density,
$\mathrm{CN}_{+s}$, and $H_{\mathrm{vap}}$. These preferences suggest
property-dependent use of the shared representations. Multiple experts
receive substantial weight for each property, allowing their representations
to contribute jointly to prediction.

We assess each expert's predictive contribution by removing it without
retraining and renormalizing the remaining routing weights.
Figure~\ref{fig:routing-mechanism}(b) shows the resulting changes in macro
z-MSE. Across the twelve equally weighted properties, removing $S_1$--$S_4$
changes macro z-MSE by +12.43\%, +6.03\%, +4.29\%, and +4.56\%, respectively.
Removing any expert increases the error, with the largest effect for $S_1$.
$S_2$ also contributes despite receiving no property's largest mean routing
weight, illustrating why allocation weights alone do not measure predictive
importance.

\FloatBarrier
\section{Conclusion}
\label{sec:conclusion}

Motivated by heterogeneous cross-property transfer, ElectrolyteFM learns private and shared representations separately and integrates them through property-specific residual adapters, enabling accurate prediction across a broad range of electrolyte properties. 
ElectrolyteFM achieves the lowest macro NMAE and the lowest MAE on 10 of the 12 properties. Conductivity OOD
and few-shot results further support its use for prediction and adaptation under formulation shift.

\clearpage

\bibliographystyle{abbrvnat}
\bibliography{main}

\appendix
\clearpage

\section{Additional Multi-Property Prediction Results}
\label{app:additional-metrics}

Tables~\ref{tab:main-pearson} and~\ref{tab:main-r2} complement the MAE
results in Table~\ref{tab:main-mae} with PCC and $R^2$ from the same test
predictions. Property order, label-source groups, and twelve-property
summaries follow Table~\ref{tab:main-mae}.

ElectrolyteFM ranks first on nine properties for each metric: all four
experimental-label targets and five computed-label targets. Its macro
PCC and $R^2$ are 0.9268 and 0.8357, respectively. 

\begin{table}[!htbp]
    \centering
    \fontsize{8}{9.4}\selectfont
    \setlength{\tabcolsep}{1.6pt}
    \setlength{\arrayrulewidth}{0.3pt}
    \renewcommand{\arraystretch}{1.08}
    \newcommand{\maehead}[3]{%
        \begingroup\fontsize{8}{9}\selectfont
        \begin{tabular}[c]{@{}c@{}}#1\\[-1pt]
        {\fontsize{7}{8}\selectfont #2}\\[-1pt]
        {\fontsize{7}{8}\selectfont #3}\end{tabular}\endgroup}
    \caption{Per-property PCC on Electrolyte12 (higher is better).
    Blocks denote experimental and computed test labels. Bold and underline
    indicate the best and second-best unrounded values. The macro and
    best-count rows summarize all twelve properties.}
    \label{tab:main-pearson}
    \begin{tabular*}{\textwidth}{@{}l|@{\extracolsep{\fill}}*{7}{c}@{}}
        \toprule
        {\fontsize{8}{9}\selectfont Property} &
        \maehead{RDKit-XGB}{(Landrum et al.}{\citeyear{landrum2024rdkit})} &
        \maehead{MolT5-XGB}{(Edwards et al.}{\citeyear{edwards2022molt5})} &
        \maehead{Bamboo-Mixer}{(Yang et al.}{\citeyear{bamboomixer2026})} &
        \maehead{GeoMix}{(Li et al.}{\citeyear{geomix2025})} &
        \maehead{MolSets}{(Zhang et al.}{\citeyear{zhang2024molsets})} &
        \maehead{SCAN}{(Wang and}{You \citeyear{wang2026scan})} &
        {\fontsize{8}{9}\selectfont\begin{tabular}[c]{@{}c@{}}\textbf{ElectrolyteFM}\\[-1pt]\textbf{(Ours)}\end{tabular}} \\
        \midrule
        \multicolumn{8}{l}{\textbf{Experimental labels}} \\
        Cond. & 0.729 & 0.748 & \underline{0.909} & 0.771 & 0.899 & 0.856 & \textbf{0.942} \\
        Visc. & 0.113 & 0.241 & 0.823 & 0.759 & \underline{0.875} & 0.258 & \textbf{0.978} \\
        $H^{E}$ & 0.419 & 0.528 & \underline{0.854} & 0.510 & 0.821 & 0.596 & \textbf{0.983} \\
        $V^{E}$ & 0.927 & 0.924 & \underline{0.955} & 0.854 & 0.919 & 0.885 & \textbf{0.959} \\
        \midrule
        \multicolumn{8}{l}{\textbf{Computed labels}} \\
        Dens. & 0.920 & 0.938 & \underline{0.997} & 0.978 & 0.986 & 0.949 & \textbf{0.999} \\
        $D_{+}$ & 0.909 & 0.904 & 0.954 & 0.958 & \underline{0.959} & \textbf{0.960} & 0.958 \\
        $D_{-}$ & 0.891 & 0.893 & 0.957 & 0.956 & 0.962 & \textbf{0.964} & \underline{0.963} \\
        $D_{s}$ & 0.889 & 0.890 & 0.966 & 0.956 & \underline{0.966} & 0.956 & \textbf{0.967} \\
        $\mathrm{CN}_{+-}$ & 0.675 & 0.677 & 0.707 & \underline{0.713} & 0.696 & 0.699 & \textbf{0.714} \\
        $\mathrm{CN}_{+s}$ & 0.664 & 0.653 & 0.678 & \underline{0.689} & 0.659 & 0.668 & \textbf{0.708} \\
        Oxid. stab. & 0.838 & 0.805 & \textbf{0.958} & 0.833 & 0.883 & 0.805 & \underline{0.953} \\
        $H_{\mathrm{vap}}$ & 0.929 & 0.896 & \underline{0.995} & 0.974 & 0.938 & 0.954 & \textbf{0.997} \\
        \midrule
        Macro PCC & 0.7418 & 0.7582 & \underline{0.8960} & 0.8294 & 0.8802 & 0.7958 & \textbf{0.9268} \\
        \textbf{Best count} & 0/12 & 0/12 & 1/12 & 0/12 & 0/12 & \underline{2/12} & \textbf{9/12} \\
        \bottomrule
    \end{tabular*}
\end{table}

\begin{table}[!htbp]
    \centering
    \fontsize{8}{9.4}\selectfont
    \setlength{\tabcolsep}{1.6pt}
    \setlength{\arrayrulewidth}{0.3pt}
    \renewcommand{\arraystretch}{1.08}
    \newcommand{\maehead}[3]{%
        \begingroup\fontsize{8}{9}\selectfont
        \begin{tabular}[c]{@{}c@{}}#1\\[-1pt]
        {\fontsize{7}{8}\selectfont #2}\\[-1pt]
        {\fontsize{7}{8}\selectfont #3}\end{tabular}\endgroup}
    \caption{Per-property $R^2$ on Electrolyte12 (higher is better).
    Blocks denote experimental and computed test labels. Bold and underline
    indicate the best and second-best unrounded values. The macro and
    best-count rows summarize all twelve properties.}
    \label{tab:main-r2}
    \begin{tabular*}{\textwidth}{@{}l|@{\extracolsep{\fill}}*{7}{c}@{}}
        \toprule
        {\fontsize{8}{9}\selectfont Property} &
        \maehead{RDKit-XGB}{(Landrum et al.}{\citeyear{landrum2024rdkit})} &
        \maehead{MolT5-XGB}{(Edwards et al.}{\citeyear{edwards2022molt5})} &
        \maehead{Bamboo-Mixer}{(Yang et al.}{\citeyear{bamboomixer2026})} &
        \maehead{GeoMix}{(Li et al.}{\citeyear{geomix2025})} &
        \maehead{MolSets}{(Zhang et al.}{\citeyear{zhang2024molsets})} &
        \maehead{SCAN}{(Wang and}{You \citeyear{wang2026scan})} &
        {\fontsize{8}{9}\selectfont\begin{tabular}[c]{@{}c@{}}\textbf{ElectrolyteFM}\\[-1pt]\textbf{(Ours)}\end{tabular}} \\
        \midrule
        \multicolumn{8}{l}{\textbf{Experimental labels}} \\
        Cond. & 0.274 & 0.303 & \underline{0.822} & 0.574 & 0.795 & 0.723 & \textbf{0.887} \\
        Visc. & 0.000 & 0.000 & \underline{0.259} & 0.025 & 0.164 & 0.005 & \textbf{0.608} \\
        $H^{E}$ & -0.129 & 0.006 & \underline{0.679} & -0.206 & 0.662 & 0.310 & \textbf{0.967} \\
        $V^{E}$ & 0.777 & 0.806 & \underline{0.910} & 0.709 & 0.824 & 0.780 & \textbf{0.915} \\
        \midrule
        \multicolumn{8}{l}{\textbf{Computed labels}} \\
        Dens. & 0.747 & 0.808 & \underline{0.995} & 0.939 & 0.972 & 0.901 & \textbf{0.998} \\
        $D_{+}$ & 0.401 & 0.490 & 0.909 & 0.877 & \textbf{0.917} & \underline{0.913} & 0.904 \\
        $D_{-}$ & 0.386 & 0.512 & 0.915 & 0.905 & \underline{0.922} & \textbf{0.927} & 0.920 \\
        $D_{s}$ & 0.487 & 0.587 & 0.906 & 0.910 & \underline{0.929} & 0.909 & \textbf{0.934} \\
        $\mathrm{CN}_{+-}$ & 0.326 & 0.351 & 0.485 & \underline{0.493} & 0.467 & 0.479 & \textbf{0.501} \\
        $\mathrm{CN}_{+s}$ & 0.282 & 0.293 & 0.441 & \underline{0.460} & 0.424 & 0.424 & \textbf{0.492} \\
        Oxid. stab. & 0.646 & 0.597 & \textbf{0.913} & 0.469 & 0.776 & 0.647 & \underline{0.908} \\
        $H_{\mathrm{vap}}$ & 0.681 & 0.621 & \underline{0.990} & 0.905 & 0.876 & 0.910 & \textbf{0.995} \\
        \midrule
        Macro $R^2$ & 0.4064 & 0.4478 & \underline{0.7687} & 0.5882 & 0.7273 & 0.6607 & \textbf{0.8357} \\
        \textbf{Best count} & 0/12 & 0/12 & \underline{1/12} & 0/12 & \underline{1/12} & \underline{1/12} & \textbf{9/12} \\
        \bottomrule
    \end{tabular*}
\end{table}

\FloatBarrier

\section{Ablation Studies and Expert Analysis}
\label{app:ablations}

\subsection{Comparison of Learning Strategies}
\label{app:learning-strategies}

Table~\ref{tab:learning-strategies-full} provides the label-source breakdown
of the twelve-property results in Table~\ref{tab:learning-strategies}.
ElectrolyteFM has the lowest macro NMAE and highest macro PCC and
$R^2$ in each group.

\begin{table}[!htbp]
    \centering\small
    \setlength{\tabcolsep}{4pt}
    \caption{Complete learning-strategy comparison. Metrics are equal-weight means over the four experimental-label or eight computed-label properties. Bold and underline indicate the best and second-best values within each column.}
    \label{tab:learning-strategies-full}
    \begin{tabular}{lrrrrrr}
        \toprule
        & \multicolumn{3}{c}{Experimental labels (4)} & \multicolumn{3}{c}{Computed labels (8)} \\
        \cmidrule(lr){2-4}\cmidrule(lr){5-7}
        Method & NMAE $\downarrow$ & PCC $\uparrow$ & $R^2$ $\uparrow$ & NMAE $\downarrow$ & PCC $\uparrow$ & $R^2$ $\uparrow$ \\
        \midrule
        STL & 0.0865 & 0.8705 & 0.7664 & \underline{0.2047} & \underline{0.9044} & \underline{0.8274} \\
        MMoE & 0.0728 & 0.9120 & 0.8214 & 0.2098 & 0.9027 & 0.8223 \\
        PLE & \underline{0.0647} & \underline{0.9647} & \underline{0.8352} & 0.2093 & 0.9019 & 0.8166 \\
        \textbf{ElectrolyteFM (Ours)} & \textbf{0.0640} & \textbf{0.9655} & \textbf{0.8442} & \textbf{0.2036} & \textbf{0.9075} & \textbf{0.8314} \\
        \bottomrule
    \end{tabular}
\end{table}

\subsection{Contributions of Knowledge Learning and Fusion}
\label{app:learning-stages}

Table~\ref{tab:learning-stages-full} reports twelve-property macro
NMAE, PCC, and $R^2$ for the configurations in
Table~\ref{tab:stage-ablation}. SpKL and ShKL
are evaluated through their own prediction heads; S2SKF
combines both learned representations. The check marks indicate included components.

\begin{table}[!htbp]
    \centering\small
    \setlength{\tabcolsep}{6pt}
    \caption{Complete knowledge learning and fusion ablation across all twelve properties.
    Check marks indicate included components; metrics give equal weight to each
    property. Bold and underline indicate the best and second-best unrounded
    values within each column.}
    \label{tab:learning-stages-full}
    \begin{tabular}{cccrrr}
        \toprule
        SpKL & ShKL &
        S2SKF & NMAE $\downarrow$ & PCC $\uparrow$ & $R^2$ $\uparrow$ \\
        \midrule
        $\checkmark$ & -- & -- & 0.1841 & 0.9171 & 0.8275 \\
        -- & $\checkmark$ & -- & \underline{0.1674} & \underline{0.9203} & \underline{0.8318} \\
        $\checkmark$ & $\checkmark$ & $\checkmark$ & \textbf{0.1571} & \textbf{0.9268} & \textbf{0.8357} \\
        \bottomrule
    \end{tabular}
\end{table}

Across all twelve properties, S2SKF reduces macro NMAE
by 14.7\% and 6.1\% relative to SpKL and ShKL, respectively. It also achieves the highest macro PCC and
$R^2$, improving aggregate performance across all three metrics.

\FloatBarrier

\subsection{Expert Allocation and Removal}
\label{app:expert-removal}

The routing weights in Figure~\ref{fig:routing-mechanism}(a) describe
which experts each property uses on average. To assess their predictive
contributions, we remove each shared expert in turn without retraining.
For expert $k$, its routing weight is set to zero and the remaining
weights are renormalized. The resulting shared state is
\[
    \mathbf{s}_{it}^{(-k)}
      =\sum_{j\ne k}\frac{\pi_{itj}}{1-\pi_{itk}}
        S_j(\mathbf{u}^{\mathrm{shr}}_i).
\]
The intervention is evaluated through the same frozen adapter and head,
using the training target transforms and standardization statistics.
\begin{table}[!htbp]
    \centering\small
    \caption{Effect of shared-expert removal on macro z-MSE (lower is better). Each column averages equally over the indicated properties; parentheses give the relative change from the complete model.}
    \label{tab:expert-removal-stratified}
    \begin{tabular}{lrrr}
        \toprule
        Intervention & Experimental (4) & Computed (8) & All (12) \\
        \midrule
        Complete model & 0.0472 & 0.1668 & 0.1269 \\
        Remove $S_1$ & 0.0790 (+67.4\%) & 0.1745 (+4.7\%) & 0.1427 (+12.4\%) \\
        Remove $S_2$ & 0.0559 (+18.4\%) & 0.1739 (+4.3\%) & 0.1346 (+6.0\%) \\
        Remove $S_3$ & 0.0534 (+13.1\%) & 0.1718 (+3.0\%) & 0.1323 (+4.3\%) \\
        Remove $S_4$ & 0.0583 (+23.4\%) & 0.1699 (+1.9\%) & 0.1327 (+4.6\%) \\
        \bottomrule
    \end{tabular}
\end{table}

Removing any expert increases twelve-property macro z-MSE, with the
largest increase for $S_1$ (+12.43\%). The effect of removing $S_1$ is much
larger for the experimental-label group (+67.35\%) than for the
computed-label group (+4.65\%). Thus average routing preference and
the cost of removal provide complementary views of expert use, and
contributions vary across the evaluated properties.

\FloatBarrier

\end{document}